\documentclass[10pt,twocolumn,letterpaper]{article}

\usepackage{cvpr}              
\definecolor{cvprblue}{rgb}{0.21,0.49,0.74}
\usepackage[pagebackref,breaklinks,colorlinks,allcolors=cvprblue]{hyperref}

\def\paperID{} 
\def\confName{CVPR}
\def\confYear{2026}

\usepackage{graphicx}	
\usepackage{amsmath}	
\usepackage{amssymb}	
\usepackage{booktabs}
\usepackage{times}
\usepackage{microtype}
\usepackage{epsfig}
\usepackage{caption}
\usepackage{float}
\usepackage{placeins}
\usepackage{color, colortbl}
\usepackage{stfloats}
\usepackage{enumitem}
\usepackage{tabularx}
\usepackage{xstring}
\usepackage{multirow}
\usepackage{xspace}
\usepackage{url}
\usepackage{bm}
\usepackage{bbm}
\usepackage{pifont}
\usepackage{makecell}
\usepackage{algorithm}
\usepackage{algpseudocode}
\usepackage{wrapfig}
\usepackage[hang,flushmargin]{footmisc}
\usepackage{graphicx}
\usepackage{ulem} 

\newcommand{\la}[1]{{#1}}

\newcommand{\lh}[1]{{#1}}

\title{MotionHiFlow: Text-to-Motion via Hierarchical Flow Matching}

\author{
  \begin{tabular}{@{}c@{}}
    Heng Li$^1$ \quad Xiaotong Lin$^1$ \quad Ling-An Zeng$^1$ \quad Yulei Kang$^1$ \quad Shuai Li$^2$ \quad Jian-Fang Hu$^{1,3,4\dagger}$ \\
    $^1$Sun Yat-sen University \qquad $^2$Shandong University \\
    $^3$ Guangdong Province Key Laboratory of Information Security Technology, China \\
    $^4$ Key Laboratory of Machine Intelligence and Advanced Computing, Ministry of Education, China \\
    [2pt]
    {\tt\scriptsize \{liheng36, linxt29, zenglan3, kangylei\}@mail2.sysu.edu.cn, shuaili@sdu.edu.cn, hujf5@mail.sysu.edu.cn}
  \end{tabular}
}

\begin{document}
\maketitle
\renewcommand{\thefootnote}{}
\footnotetext{$^{\dagger}$ Corresponding Author}


\begin{abstract}
    Text-to-motion generation aims to generate 3D human motions that are tightly aligned with the input text while remaining physically plausible and rich in fine-grained detail. Although recent approaches can produce complex and natural movements, they usually operate at only one temporal scale, which limits both semantic alignment and temporal coherence. Inspired by the fact that complex motions are conceptualized hierarchically rather than at a single temporal scale in the human cognitive system, we propose \textit{MotionHiFlow}, a hierarchical flow matching framework to generate motion progressively by constructing flow path from low to high temporal scales. The flows at lower scales capture high-level semantics and coarse motion structures, while flows at higher scales refine temporal details. To link the flows across scales, we introduce a novel cross-scale transition process, ensuring continuity and preserving noise consistency. Furthermore, by integrating a Text-Motion Diffusion Transformer and a topology-aware Motion VAE, MotionHiFlow explicitly models structural dependencies among joints via joint-aware positional encoding and skeletal topology, enabling precise semantic alignment alongside fine-grained motion details. Extensive experiments on HumanML3D and KIT-ML benchmarks demonstrate state-of-the-art performance, with ablation studies confirming the effectiveness of the hierarchical design and key components. Code is available at \url{https://github.com/ai-lh/MotionHiFlow}.
\end{abstract}    
\section{Introduction}
\label{sec:intro}

Text-to-motion generation aims to synthesize realistic 3D human motions conditioned on natural language descriptions, with broad applications in virtual reality, character animation, and robotics. The generated motions are expected to be semantically consistent with the input text and physically plausible with fine-grained motion details. Benefiting from advances in generative modeling and powerful sequence learning architectures, recent methods~\cite{Momask,BAMM,MoGenTS,MARDM,Morph,TAAT} have made notable progress in generating complex and natural human motions at a single temporal scale.

In the human cognitive system ~\cite{DBLP:conf/cvpr/MaNLZL22}, complex motions are conceptualized hierarchically rather than at a single temporal scale. Humans typically realize human motions by first constructing a high-level framework of key poses (termed as coarse motion), and subsequently refine it with dynamic transitions and fine-grained limb movements to produce coherent fine motion. However, different from such coarse-to-fine cognitive process, recent methods~\cite{Momask,BAMM,MoGenTS} simultaneously model semantic alignment and motion details at a single temporal scale, which limits their ability to achieve long-term coherence, naturalness, and precise alignment with textual input. To address this gap, we {aim to design} a hierarchical coarse-to-fine generation strategy that first produces a coarse motion to capture high-level semantic structure at a low temporal \lh{scale}, and then progressively refines it by adding fine-grained motion details at higher temporal scales.




\begin{figure}[t]
    \centering
    \includegraphics[width=\linewidth]{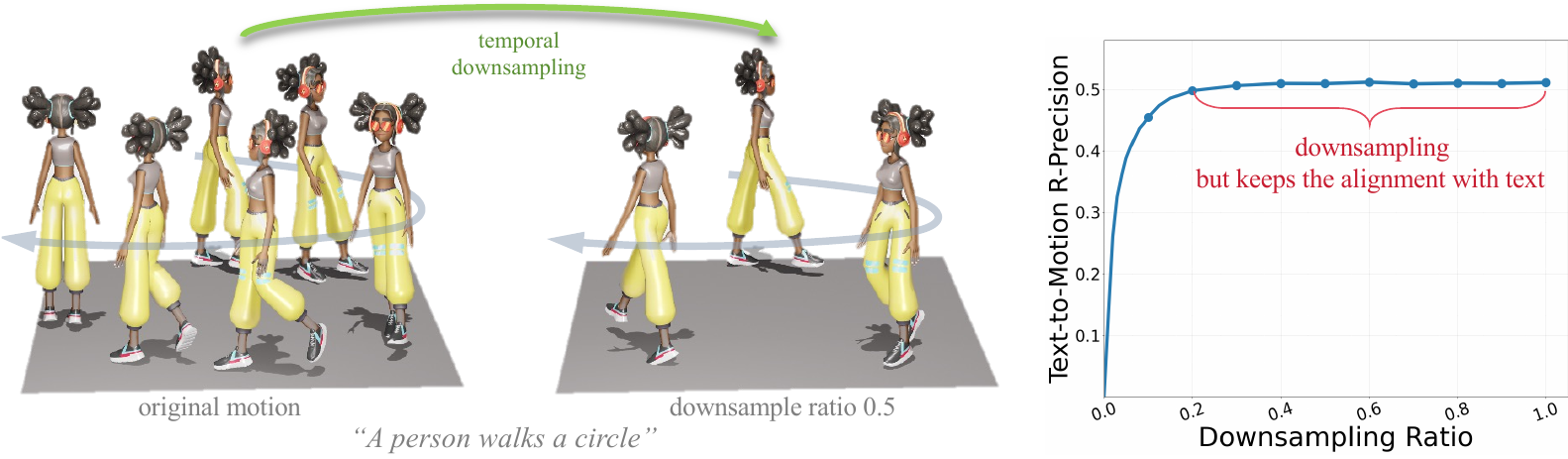}
    \caption{
    Text-to-Motion retrieval precision under different downsampling ratios.
    The R-precision remains stable as the down-sampling ratio decrease, which means that models trained on coarse motions can achieve robust semantic alignment.
    }
    \label{fig:motivation}
    \vspace{-0.4cm}
\end{figure}

To verify this intuition, we first evaluate how much semantic information is preserved when motion is viewed at a lower temporal scale (\Cref{fig:motivation}). Starting from the original motion, we generate coarse motions by linearly downsampling and evaluate text–motion alignment with the R-precision metric on the HumanML3D \cite{humanml3d} test set.  Remarkably, R-precision remains stable as the downsampling ratio decreases even at 0.2× (i.e., retaining only 20\% of the frames), indicating that coarse motions preserve most of the semantics described in the text. Moreover, by training motion generation models at different scales \lh{(refer to Table~\ref{tab:HT})}, we observe that models trained solely on coarse motions often achieve robust semantic alignment, sometimes even outperforming those trained on fine-scale motions. These observations suggest that overemphasizing fine-grained details may hinder semantic learning, while training on coarse motions promotes stronger alignment with the core textual semantics.

Building on the above observations, we propose a hierarchical framework \lh{\textit{MotionHiFlow}} to generate motion progressively from low to high temporal scales across multiple stages, aiming to achieve both strong semantic alignment and rich motion details. 
Specifically, we devise a stage-wise flow to link the start (noisier latent) and end (cleaner latent) at each scale. 
Instead of directly upsampling lower-scale noisy data as done in other works~\cite{pyramidal-flow, chen2025pixelflow}, here we formulate a novel cross-scale transition process to link flows across scales, which contains: 1) denoising: constructing the clean data at the lower scale by extrapolation; 2) upsampling: generating clean data at higher scale with upsampling; 3) renoising: constructing the noise data at the higher scale via interpolation.
\lh{We define the transition in this way such that noise consistency can be preserved across stages.}
By integrating the cross-scale transition with stage-wise flows, a generative process is established that maps noise to data.
To operationalize \lh{MotionHiFlow}, we further introduce \lh{\textit{Text-Motion Diffusion Transformer (TMDiT)}}, a novel motion generation {model} that \lh{harnesses} hierarchical flow matching for efficient and smooth motion generation.
\lh{Built upon recent success of diffusion models in image generation~\cite{StableDiffusion3,DiT,flux2024},} TMDiT explicitly incorporates the inherent structural dependencies among human joints through joint-aware positional encoding (Joint RoPE). 
By combining TMDiT with \lh{{a} topology-aware Motion VAE} \lh{which encodes motion sequence into latent}, our model effectively generates motions with both semantic alignment and fine-grained details. 





We conduct quantitative and qualitative comparisons together with extensive ablation studies on the HumanML3D \cite{humanml3d} and KIT-ML \cite{kit-ml} datasets to verify the effectiveness of our hierarchical design and each model component. 
In summary, our main contributions are as follows: 
\begin{itemize}[leftmargin=1em]
\item {We propose \textit{MotionHiFlow}, a hierarchical flow matching framework for text-to-motion generation that progressively and consistently generates motion from low to high temporal scales, achieving strong semantic alignment and rich fine-grained motion details.}
\item We develop the \textit{Text-Motion Diffusion Transformer} (TMDiT), which explicitly incorporates the inherent structural dependencies among human joints through joint-aware positional encoding and skeletal topology. 
\item We achieve state-of-the-art performance on the HumanML3D and KIT-ML datasets, with comprehensive ablations validating the effectiveness of our approach.
\end{itemize}

\section{Related work}
\label{sec:related}

\subsection{Text-Conditioned Motion Generation} 
{Text-conditioned human motion generation is a challenging task. Early attempts~\cite{ahuja2019language2pose, humanml3d, TEMOS} primarily focused on learning direct mappings or shared embeddings between text descriptions and motions. For instance, Language2Pose~\cite{ahuja2019language2pose} proposed learning a common latent representation subspace for both modalities. Despite significant progress, these methods still struggle to generate diverse and high-fidelity motions. More recently, diffusion-like models~\cite{MDM, Motiondiffuse, MLD, CrossDiff, MMM, MoGenTS, light-t2m, fg-t2m, mofusion, motionlcm, sampieri2024length, emdm, DBLP:conf/icmcs/LiLLZH25} have become the dominant approach for text-conditioned human motion generation, significantly advancing the field. For example, MoGenTS~\cite{MoGenTS} builds on this by enhancing the spatial VAE component for improved motion representation. Furthermore, other methods include autoregressive approaches~\cite{t2m-gpt,tm2t, MotionGPT, chen2024motionllm, avatargpt}, generative masked modeling techniques~\cite{Momask, BAMM, zhang2025towards}, and additional types~\cite{bridgegap, humantomato, wandr, parco, attt2m,motionagent,MotionStreamer,ScaMo,li2025irg,DBLP:conf/icmcs/HuangZZGZ25,ChainHOI,tang_pami,sun2022you,sun2021action}. Moreover, some works~\cite{wandr, jin2024local, pmg, tlcontrol, omnicontrol, gmd, kong2023priority,maskcontrol,PersonaBooth} explore using supplementary conditions to control the generated motion. However, these models operate at a single temporal scale, which limits their ability to simultaneously capture global trajectory structures (requiring coarse temporal views) and fine-grained motion details (necessitating fine temporal views). In contrast, we develop a hierarchical flow matching framework that enables multi-scale generation, starting with high-level semantic alignment to the text and progressively refining motion details in a top-down manner.}

\vspace{-0.1cm}
\subsection{Flow Generative Models} \vspace{-0.1cm}
Flow Matching, a powerful generative modeling paradigm~\cite{flow-matching,Rectified-Flow}, has recently achieved compelling results in domains including image and video generation~\cite{StableDiffusion3,flux2024,pyramidal-flow}. By leveraging theoretical advantages such as straight flow formulations and direct conditional path probability learning, flow generative models offer benefits in training stability and efficient generation compared to diffusion-based models. For example, Patrick et al.~\cite{StableDiffusion3} propose to learn a continuous path between a high-dimensional simple noise distribution and the target high-resolution image manifold. 
However, flow generative models in 3D human motion generation remain largely underexplored. Although a few methods have made initial attempts~\cite{flowmotion, motion-flow}, they naively apply flow matching for generation in the motion space without adaptation, resulting in suboptimal utilization and generation performance. To address this, we enhance flow matching by integrating Text-Motion Diffusion Transformer with a joint-aware positional encoding, fully unlocking the significant potential of {flow matching models} for human motion generation.

\section{Method}
\label{sec:method}

\begin{figure*}[t]
    \centering
    \includegraphics[width=0.99\textwidth,height=0.38\textwidth]{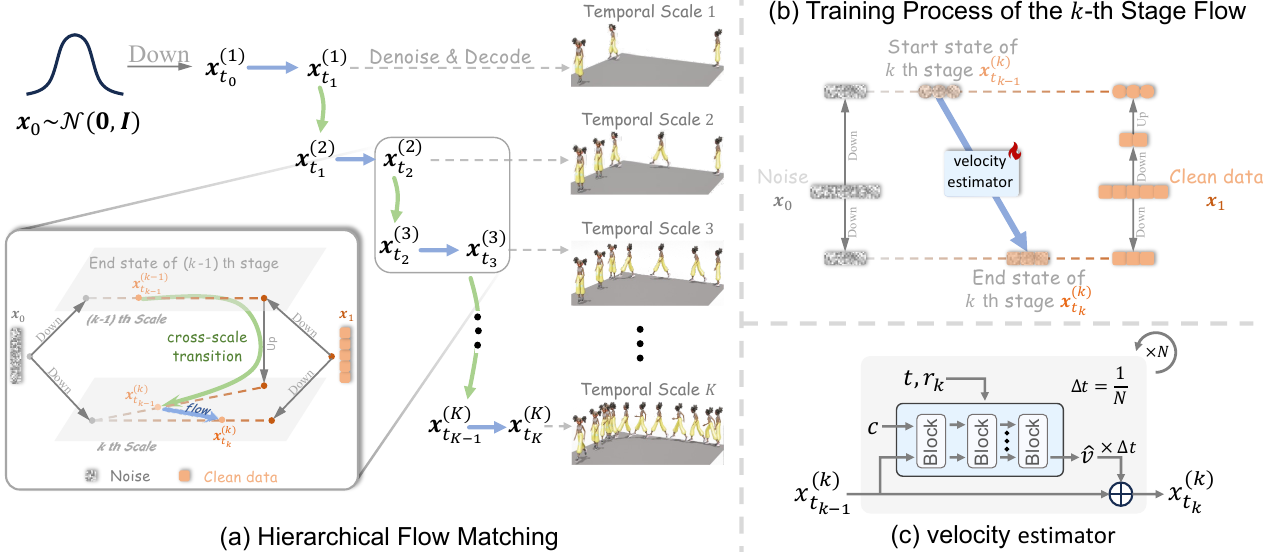}
    \vspace{-0.15cm}
    \caption{Overview of our MotionHiFlow, which progressively generates motion from low to high temporal scales across multiple stages. The early stages mainly capture high-level semantics and coarse motion structures, while later stages model fine-grained temporal details via \textit{cross-scale transition} and \textit{flow} operators. \lh{The points along the gradient-colored dashed line in the inset (bottom left) of (a) and in (b) denotes a linear interpolation between its endpoints.} \lh{Down/Up denotes downsampling/upsampling, respectively.}}
    \vspace{-0.13cm}
    \label{fig:HFM}
\end{figure*}

\subsection{Problem Formulation}
Given a text query describing human motion or action, our goal is to generate a corresponding 3D human pose sequence $M = \{m_i\}$ of length $L$. According to existing works~\cite{humanml3d, Momask}, each joint in the 3D human pose is represented by the root angular velocity along the Y-axis, root linear velocities on the XZ plane, root height, and local joint positions, rotations and velocities relative to the root space. Please refer to T2M~\cite{humanml3d} for more detailed information on human pose representation. 
Each human pose is represented as a $J \times D_j$-shaped tensor, where $J$ indicates the joint number, and $D_j$ is the dimension of joint representation.

\subsection{Overview}
To generate semantically aligned, temporally coherent, and detailed motion, we {propose a novel hierarchical flow matching framework, MotionHiFlow.
This framework operates in the latent space \lh{encoded} by \lh{a} Motion VAE that {provides} topology-aware encoding of motion sequences. {As shown in \Cref{fig:HFM},} our framework progressively generates the motion from low to high temporal scales across multiple stages.} The early stages mainly capture high-level semantics and coarse motion structures, while later stages add fine-grained temporal details. 
\lh{To facilitate the continuity flow matching across stages, we formulate a novel cross-scale transition with denoising-upsampling-renoising process, preserving noise consistency over the entire motion generation process.}
\lh{By integrating the cross-scale transition with stage-wise flows, a generative process is established that maps noise to data.}
{Other} key components include \lh{a} Motion VAE with a \lh{two-stream} Graph Convolutional Networks~\cite{2s-agcn} for motion tokenization, a Text-Motion Diffusion Transformer (TMDiT) modeling hierarchical flow conditioned on text, and a Joint RoPE mechanism for improved positional encoding. \lh{Combining with these components, the clean latents outputted by our} MotionHiFlow can be decoded to realistic, expressive motions that closely match textual descriptions.


\subsection{Preliminaries: Flow Matching} 
Flow matching models~\cite{flow-matching,Rectified-Flow, StableDiffusion3} aim to learn a \lh{velocity} field $u_t$ that transforms noise $\bm{x}_0 \sim \mathcal{N}(\bm{0}, \bm{I})$ {into} data samples \(\bm{x}_1 \sim q_{\text{data}}\). To achieve this, a neural network \(v_\theta(\bm{x}_t,t)\) parameterized by \(\theta\), is trained to approximate the target field \(u_t(\bm{x}_t|\bm{x}_1)
\) by minimizing the following loss:
\begin{equation}
    \mathcal{L}_{FM}(\theta)=\underset{\substack{t\sim\mathcal{U}(0,1) \\ p_t(\bm{x}_t|\bm{x}_1), q(\bm{x}_1)}}{\mathbb{E}}\|v_\theta(\bm{x}_t,t)-u_t(\bm{x}_t|\bm{x}_1)\|^2. \label{eq:cfm_loss}
\end{equation}
Here, $p_t(\bm{x}_t|\bm{x}_1)$ defines the conditional probability path linking intermediate point $\bm{x}_t$ to $\bm{x}_1$. A common choice~\cite{flow-matching} for $\bm{x}_t$ at this path is linear \lh{interpolation} between $\bm{x}_0$ and $\bm{x}_1$:
\begin{equation}
    \bm{x}_t=(1-t)\bm{x}_0+t\bm{x}_1.\label{eq:def_x_t}
\end{equation}
Then, the corresponding target \lh{velocity} field simplifies to $u_t(\bm{x}_t|\bm{x}_1)=\bm{x}_1-\bm{x}_0$. The new data can be constructed by  solving the following \lh{Ordinary Differential Equation (ODE)} with existing \lh{solvers} (e.g., Euler, RK45~\cite{RK45}):
\begin{equation}
\mathrm{d}\bm{x}_t=v_\theta(\bm{x}_t,t)\mathrm{d}t. \label{eq:flow_ode}
\end{equation}

\subsection{Hierarchical Flow Matching {Framework}} \label{subsec:HFM}
We propose a hierarchical flow matching framework for \la{the} text-to-motion \la{generation task}, which \la{generates motion progressively from low to high temporal scales across multiple stages, aiming to achieve both strong semantic alignment, high temporally coherent and rich detailed motion.} As illustrated in \Cref{fig:HFM} (a), {our framework} contains \(K\) stages of generation, each operates at a certain temporal scales.
Early stages {focus on the semantics} of coarse motion and text alignment, and the subsequent stages intend to progressively enrich the details of the motion provided by preceding stage.

Specifically, in the $k$-th stage, our method processes motion at temporal scale \(r_k \in (0,1]\) {and refines} the motion representation within time interval \([t_{k-1}, t_k]\), where \(\{t_k\}_{k=0}^K\) are time points partitioning the interval \([0,1]\).
We utilize flow matching to learn a \lh{flow} transformation \(S_k\) that maps the start state $\bm{x}_{t_{k-1}}^{(k)}$ to end state $\bm{x}_{t_k}^{(k)}$, {as defined below}:
\begin{align}
    &\text{Start:}
        &&\bm{x}_{t_{k-1}}^{(k)} = \begin{aligned}[t] 
                       & (1-t_{k-1})f(\bm{x}_0, r_k) \\
                       & + t_{k-1} f(f(\bm{x}_1, r_{k-1}), r_k/r_{k-1})
                   \end{aligned} \label{eq:start_k}, \\
    &\text{End:} 
        &&\bm{x}_{t_k}^{(k)} = (1-t_k)f(\bm{x}_0, r_k) + t_k f(\bm{x}_1, r_k).
           \label{eq:end_k}
\end{align}
Here, $f(\bm{x},r)$ means performing \lh{a} temporal \lh{resampling} \lh{on} $\bm{x}$ \lh{with} a factor $r$, it is downsampling when $r\in(0,1)$ and upsampling when $r>1$. The end state $\bm{x}_{t_k}^{(k)}$ \lh{at scale $r_k$ is defined as a} linear \lh{interpolation} between noise $\bm{x}_0$ and clean data $\bm{x}_1$. The start state $\bm{x}_{t_{k-1}}^{(k)}$ \lh{is carefully designed to incorporate the information $f(\bm{x}_1, r_{k-1})$ from the previous stage and initial noise $\bm{x}_0$, maintaining noise consistency across stages}.
\lh{By integrating the flow \(S_k\)
 across stages, we define the generative path of MotionHiFlow from noise to data, which can be} trained by minimizing the following loss:
\begin{equation}
    \mathcal{L}_{HFM}(\theta)=\underset{\substack{k,t}}{\mathbb{E}}\left\|v_{\theta}(\bm{x}_t^{(k)}, t) - (\bm{x}_{t_k}^{(k)} - \bm{x}_{t_{k-1}}^{(k)})\right\|^2. \label{eq:HFM-loss}
\end{equation}
Here, $\bm{x}_{t}^{(k)}$ represents training points sampled in $k$ $\text{th}$ stage, which is defined as:
\begin{equation}
    \bm{x}_{t}^{(k)} = (1-\tau)\bm{x}_{t_{k-1}}^{(k)} +\tau \bm{x}_{t_k}^{(k)}, \label{eq:hiflow_xt}
\end{equation}
where \(\tau = (t-t_{k-1})/(t_k-t_{k-1})\) is the normalized time within the stage {\( k\)}.

\lh{During inference, }
instead of directly upsampling lower-scale noisy data as done in other works~\cite{pyramidal-flow, chen2025pixelflow}, 
\lh{which leads to noise inconsistency and thus degrades generation performance}.
Here we formulate a novel cross-scale transition process to bridge the start state of stage $k+1$ with the end state of stage $k$ across scales. \lh{This process} involves three steps: 1) denoise: constructing the clean data at the lower scale by extrapolation; 2) upsample: generating clean data at the higher scale with upsampling; 3) renoise: constructing the noise data at the higher scale via interpolation.
Formally, the transition is governed by the following equations:
\begin{align}
    &\text{denoise:}
        && \hat{\bm{x}}_{1}^{(k)} = \left[\hat{\bm{x}}_{t_k}^{(k)} - (1 - t_k) \bm{x}_0^{(k)}\right]/{t_k}, \label{eq:jump1} \\
    &\text{upsample:}
        &&\hat{\bm{x}}'^{(k+1)}_1 = f(\hat{\bm{x}}_{1}^{(k)}, {r_{k+1}}/{r_k}), \label{eq:jump} \\
    &\text{renoise:} 
        &&\hat{\bm{x}}_{t_k}^{(k+1)} = (1 - t_k) \bm{x}_0^{(k+1)} + t_k\hat{\bm{x}}'^{(k+1)}_1. \label{eq:jump2}
\end{align}
Here, $\bm{x}_0^{(k)}$ denotes $f(\bm{x}_0, r_k)$, and similarly for $\bm{x}_0^{(k+1)}$.
This three-step process preserves noise consistency and ensures smooth transitions across scales, thus facilitating robust inference across all stages.
The complete inference process is presented in Algorithm~\ref{alg:inference}. 

By integrating the cross-scale transition with stage-wise flows, 
this generative process constitutes a deterministic ODE trajectory from the initial noise $\bm{x}_0$ towards the data distribution, avoiding the need to introduce additional noise between stages as done in related methods~\cite{pyramidal-flow, chen2025pixelflow}.

\begin{algorithm}[tbp]
\caption{Hierarchical Flow Matching Inference}
\label{alg:inference}
\begin{algorithmic}[1] 
    \Require Trained model parameters \(\theta\) (for \(v_{\theta}\)), scale schedule \(r_1, \dots, r_K\), time partition intervals \([t_0, t_1, \dots, t_{K-1}, t_K]\) with \(t_0=0, t_K=1\), temporal linear resampling function \(f(\cdot, \cdot)\), prior distribution for noise (e.g., \(\mathcal{N}(\bm{0}, \bm{I})\)), conditioning \(c\).

    \Ensure Generated \lh{clean} sample \(\hat{\bm{x}}_1\) at \lh{full} scale.

    \State Sample {the} initial noise $\bm{x}_0 \sim \mathcal{N}(\bm{0}, \bm{I})$
    \State Initialize {the} start state: $\hat{\bm{x}}_{0}^{(k)} \leftarrow f(\bm{x}_0, r_1)$

    \For{$k = 1$ to $K$}
        \State \lh{Flow} from $\hat{\bm{x}}_{t_{k-1}}^{(k)}$ to $\hat{\bm{x}}_{t_k}^{(k)}$
        \Comment{Equation~\eqref{eq:flow_ode}}
        \State Cross-scale transition, get {the} start state of {the} next scale $\hat{\bm{x}}_{t_k}^{(k+1)}$
        \Comment{Equation~\eqref{eq:jump1}, \eqref{eq:jump} and \eqref{eq:jump2}}
    \EndFor

    \State $\hat{\bm{x}}_1 \leftarrow f({\hat{\bm{x}}}_{t_K}^{(K)}, 1/r_K)$
    \Comment{$r_K$ may be smaller than 1}

    \State \Return $\hat{\bm{x}}_1$
\end{algorithmic}
\end{algorithm}

\begin{figure*}[t]
    \centering
    \includegraphics[width=0.9\textwidth,height=0.33\textwidth]{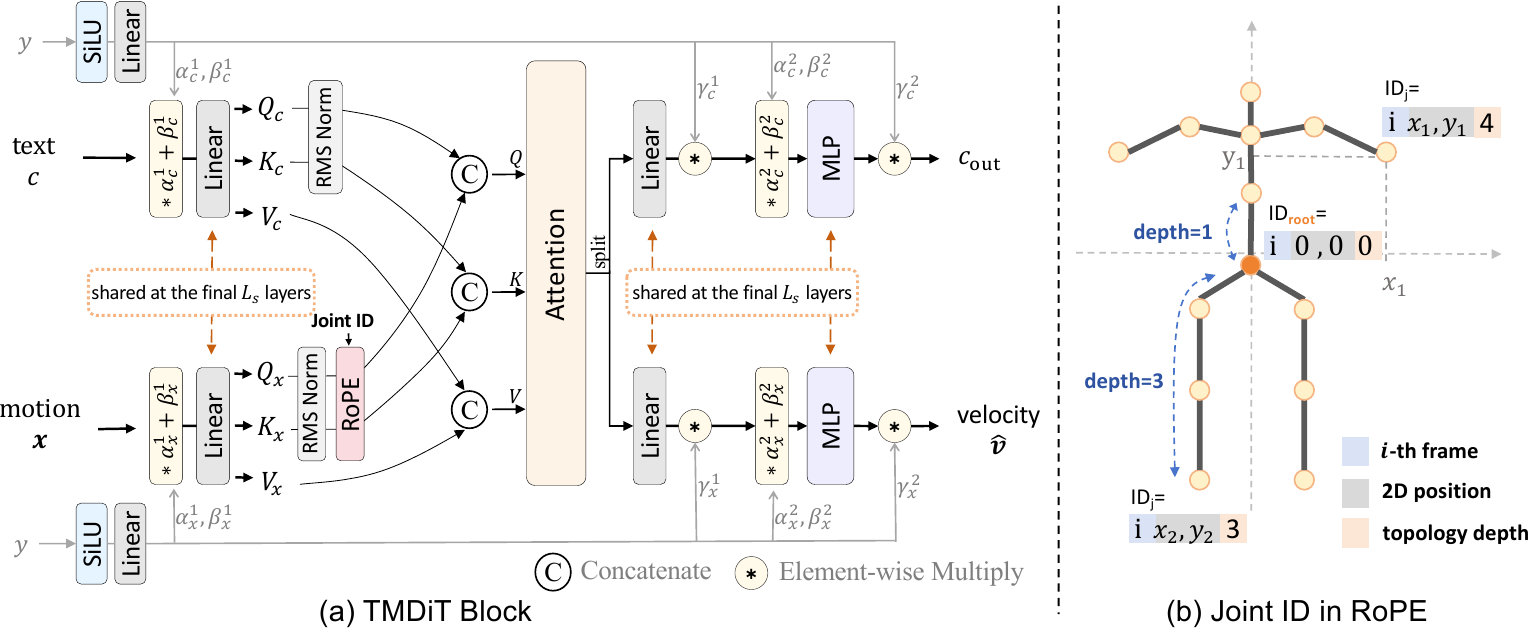}
    \vspace{-0.15cm}
    \caption{{Illustration of two main components in our TMDiT. (a) The TMDiT block employs two separate streams that independently process motion and text features, while \lh{self}-attention and shared parameters enables information exchange between streams. (b) The Joint RoPE integrates rotations derived from temporal displacement, relative spatial coordinates, and the human body topologies. Here, $x_j$ and $y_j$  denote the $j$-th joint's coordinates in the $xy$-plane. }}
    \label{fig:Arch}
    \vspace{-0.05cm}
\end{figure*}

\subsection{Model Architecture} \label{sec:architecture}
{For efficient and stable training, we introduce a specific model architecture comprising several key elements.}
First, we employ a \lh{Motion VAE} to project input motion into a \lh{topology-aware} latent space and then reconstruct the motions from latent representation.
Then, we develop the Text-Motion Diffusion Transformer (TMDiT), which explicitly incorporates the inherent structural dependencies among human joints through joint-aware positional encoding (Joint RoPE). These components combined with MotionHiFlow form a strong text-to-motion generation model.


\vspace{0.1cm}
\noindent
\textbf{Motion VAE.}
Our Motion VAE is utilized to encode the input motion $M \in \mathcal{R}^{L\times J \times D_j}$ into a compact latent representation $\bm{x}\in\mathcal{R}^{l\times j\times d}$ and reconstruct the original motion sequence from this compact latent representation. 
{Here, $L$, $J$, and $D_j$ denote the number of frames, the number of input joints per frame, and the dimension of joint features (e.g., rotation, velocity), respectively.}
Correspondingly, $l$ is the latent temporal length and $j$ is the number of latent joints.
Unlike MoGenTS~\cite{MoGenTS}, which primarily employs 2D convolutions, our {Motion VAE} utilizes Graph Convolutional Networks (GCNs)~\cite{2s-agcn,graph-pool,hong2025salad} to explicitly capture human body topology. The encoder performs temporal downsampling by a factor \(4\) (resulting in \( l = \lfloor L/4 \rfloor)\)), and spatial graph downsampling (from $J$ to $j$ latent joints) using methods such as averaging or learnable pooling~\cite{graph-pool}. This topology-aware encoding yields modest improvements in reconstruction quality, it notably enhances generation quality when combined with our TMDiT and Joint RoPE.

\vspace{0.1cm}
\noindent
\textbf{Text-to-Motion Diffusion Transformer (TMDiT).}
We present the Text-to-Motion Diffusion Transformer (TMDiT), a new architecture for generating human motion conditioned on textual descriptions. Our approach is motivated by recent progress in diffusion transformers, notably MMDiT~\cite{StableDiffusion3} and Flux~\cite{flux2024}. Crucially, TMDiT departs from conventional methods (e.g., \cite{Momask,MMM,MoGenTS}) that typically adapt vanilla Transformers~\cite{Transformer} by processing motion sequences alongside a single, sentence-level text embedding ${c}_{\text{vec}}$. We would like to point out that this prior strategy limits the nuanced interplay required between text and motion.
As illustrated in Figure~\ref{fig:HFM} (b), TMDiT receives noised motion latent $\bm{x}_{t}^{(k)}$and conditioning word-level text embedding $\bm{c}$ encoded by CLIP~\cite{CLIP}. Information about the current timestep $t$, sentence-level text embedding $c_{\text{vec}}$ and staged scale $r_k$ is fused into an embedding $\bm{y}$, which then modulates the TMDiT blocks~\cite{StableDiffusion3}. The output corresponding to the motion component serves as the estimated \lh{velocity} field for the underlying ODE solver.

Within a TMDiT block, as shown in Figure~\ref{fig:Arch} (a), we employ separate processing streams for the motion features $\bm{x}$ and text features $\bm{c}$. Both $\bm{x}$ and $\bm{c}$ are separately fed into distinct linear transformations before and after the attention, as well as within the feedforward MLP. The conditioning embedding $\bm{y}$ modulates these operations through scaling and shifting (pre-attention and pre-MLP) and gating (post-attention and post-MLP), akin to established practices in diffusion models~\cite{StableDiffusion3,DiT}. 
\lh{For clarity of presentation in Figure~\ref{fig:Arch} (a), LayerNorm operations are not explicitly shown, which precede the scaling/shifting.}
To effectively capture shared representations while preserving modality-specific nuances, TMDiT employs a parameter sharing scheme~\cite{flux2024}. The \lh{early} layers of the network utilize separate parameters for motion and text pathways, allowing for independent feature extraction. Conversely, the final $L_s$ layers share parameters. It first refines modality-specific characteristics and then encourages the learning of common representations essential for coherent text-to-motion generation.


\vspace{0.1cm}
\noindent
\textbf{Joint RoPE.}
We introduce Joint RoPE, an adaptation of Rotary Position Embedding~\cite{RoFormer} optimized for skeletal motion generation. Joint RoPE encodes relative positions by integrating rotations derived from temporal displacement, relative spatial coordinates, and the kinematic tree structure. A key innovation is its enforcement of skeletal symmetry, ensuring identical relative rotations for symmetrically equivalent joint pairs (e.g., left-hand to right-hand vs. left-foot to right-foot) at the same temporal offset, embedding structural bias akin to RoPE’s temporal encoding. Within each attention head, feature dimensions are divided into four segments with proportions \([1/2, 1/8, 1/8, 1/4]\), each applying a 1D RoPE. The first segment \((1/2)\) encodes the joint’s temporal position in the motion sequence. The next two segments \((1/8\) each, totaling \(1/4\)) encode the joint’s 2D spatial coordinates relative to the pelvis in a reference T-pose. The final segment \((1/4\)) encodes the joint’s depth in the kinematic tree, with the pelvis as the root. Temporal indices are scaled by a factor \(r_k\) before applying RoPE, enabling the TMDiT to generate spatially and structurally coherent motions. By unifying all positional encodings under RoPE framework, Joint RoPE 
can yield performance improvements, offer potential scalability to varying joint counts, and enhance adaptability across diverse skeletal structures.

\subsection{Training and Inference}  \label{subsec:train_inference}  
\vspace{0.1cm}
\noindent
\textbf{Two-Stage Training.}
{Our training pipeline has two stages. 
\lh{In the first stage}, we train the Motion VAE by minimizing a composite objective that combines the standard VAE losses (reconstruction and KL loss) and an auxiliary term that improves the temporal robustness of the latent representation. 
For a random subset of each batch, we downsample the latent vector $\bm{x}$ by a factor $r \in [0.3,1]$ with a linear \lh{resampling} function $f$, then compute the mean-squared error (MSE) between the Motion VAE decoder output (generated from the downsampled latent) and the corresponding downsampled motion sequence $M$:
}
\begin{equation}
    \mathcal{L}_{\text{aug}} = \|\text{Dec}(f(x,r)) - f(M,r)\|^2. \label{eq:loss_aug}
\end{equation}
In the second stage, we freeze the Motion VAE and train the TMDiT model $v_\theta$ using the hierarchical flow-matching loss (Eq.~\ref{eq:HFM-loss}).
To enable classifier-free guidance (CFG) \cite{CFG}, we apply condition dropout during training, randomly replacing the text condition $c$ with a null token $\varnothing$ at 10 \% probability. Training is conducted following the standard flow matching paradigm for each scale $r_k$, with the timestep $t$ uniformly sampled from the interval $[t_{k-1},\,t_k]$.

\vspace{0.1cm}
\noindent
\textbf{Inference.}
\lh{Given the well-trained model parameters ${v}_\theta$ and random noise \({\bm{x}}_0\), we generate the motion latent representations  \(\hat{\bm{x}}_1\) following the procedure presented in Algorithm~\ref{alg:inference}.}
\lh{Specifically, we enhance the velocity estimation with CFG~\cite{CFG}. The resulting latent  \(\hat{\bm{x}}_1\) is finally inputted to the Motion VAE decoder, producing the motion sequence.}

\section{Experiments}
\label{sec:exp}

\begin{table*}[t]
\caption{Quantitative comparisons with the current state-of-the-art methods on the HumanML3D (upper half) and KIT-ML (lower half) datasets. Symbol ``±" denotes a \(95\%\) confidence interval. Text in \textbf{bold} and \underline{underline} denote the best and \underline{second-best} results, respectively.}
\vspace{-0.3cm}
\small 
\begin{center}
\setlength{\aboverulesep}{0.2ex}
\setlength{\belowrulesep}{0.3ex}
{
\begin{tabular}{llcccccc}
\toprule[1.5pt]
\multirow{2}{*}{{Methods}} & \multirow{2}{*}{{Venue}} & \multicolumn{3}{c}{{R-Precision} $\uparrow$} & \multirow{2}{*}{{FID}$\downarrow$}  & \multirow{2}{*}{\makecell[c]{MultiModal\\ Dist} $\downarrow$} 
& \multirow{2}{*}{\makecell[c]{Diversity \\ $\rightarrow$} } \\
\cmidrule{3-5}
&& {Top1} & {Top2} & {Top3} & &  \\
\midrule[1.5pt]
\rowcolor{lightgray} \multicolumn{8}{l}{\textit{On the HumanML3D dataset \la{\cite{humanml3d}}.}} \\
\midrule[0.5pt]
     TEMOS \cite{TEMOS} & \textit{ECCV}'22  & $0.424^{\pm.002}$ & $0.612^{\pm.002}$ & $0.722^{\pm.002}$ & $3.734^{\pm.028}$ & $3.703^{\pm.008}$ & $8.973^{\pm.071}$\\
     T2M-GPT \cite{t2m-gpt} & \textit{CVPR}'23 & $0.492^{\pm.003}$ & $0.679^{\pm.002}$ & $0.775^{\pm.002}$ & $0.141^{\pm.005}$ & $3.121^{\pm.009}$ & $9.761^{\pm.081}$ \\
     ReMoDiffuse \cite{zhang2023remodiffuse} & \textit{ICCV}'23 & $0.510^{\pm.005}$ & $0.698^{\pm.006}$ & $0.795^{\pm.004}$ & $0.103^{\pm.004}$ & $2.974^{\pm.016}$ & $ 9.018^{\pm.075}$ \\
     MoMask \cite{Momask}  & \textit{CVPR}'24 & $0.521^{\pm.002}$ & $0.713^{\pm.002}$ & $0.807^{\pm.002}$ & ${0.045}^{\pm.002}$ & $2.958^{\pm.008}$ & - \\
     BAMM \cite{BAMM}  & \textit{ECCV}'24 & ${0.525}^{\pm.002}$ & ${0.720}^{\pm.003}$ & ${0.814}^{\pm.003}$ & $0.055^{\pm.002}$ & $2.919^{\pm.008}$ & $9.717^{\pm.089}$ \\
     MoGenTS \cite{MoGenTS}  & \textit{NeurIPS}'24 & ${0.529}^{\pm.003}$ & ${0.719}^{\pm.002}$ & ${0.812}^{\pm.002}$ & $\underline{0.033}^{\pm.001}$ & ${2.867}^{\pm.006}$ & $9.570^{\pm.077}$ \\
     Light-T2M \cite{light-t2m} & \textit{AAAI}'25 & $0.511^{\pm.003}$ & $0.699^{\pm.002}$ & $0.795^{\pm.002}$ & $0.040^{\pm.002}$ & $3.002^{\pm.008}$ & -\\
     IRG-MotionLLM \cite{li2025irg} & \textit{arXiv}'25 & $0.535^{\pm.002}$ & $0.725^{\pm.002}$ & $0.820^{\pm.002}$ & $0.242^{\pm.006}$ & $2.785^{\pm.006}$ & $9.900^{\pm.094}$\\
     EnergyMoGen \cite{EnergyMoGen} & \textit{CVPR}'25 & $0.526^{\pm.003}$ & $0.718^{\pm.003}$ & $0.815^{\pm.002}$ & $0.176^{\pm.006}$ & $2.931^{\pm.007}$ & $9.500^{\pm.091}$ \\
     SALAD \cite{hong2025salad} & \textit{CVPR}'25 & $\bm{0.581}^{\pm.003}$ & $\bm{0.769}^{\pm.003}$ & $\bm{0.857}^{\pm.002}$ & $0.076^{\pm.002}$ & $\bm{2.649}^{\pm.009}$ & $9.696^{\pm.096}$ \\
     
     MoMask++ \cite{guo2025snapmogen} & \textit{NeurIPS}'25 & ${0.528}^{\pm.003}$ & ${0.718}^{\pm.003}$ & ${0.811}^{\pm.002}$ & ${0.072}^{\pm.003}$ & ${2.912}^{\pm.008}$ & - \\
    \midrule
     MotionHiFlow (ours) & - & $\underline{0.563}^{\pm.003}$ & $\underline{0.754}^{\pm.003}$ & $\underline{0.843}^{\pm.003}$ & $\bm{0.032}^{\pm.002}$ & $\underline{2.691}^{\pm.009}$ & $9.504^{\pm.071}$ \\
\midrule[1.5pt]
\rowcolor{lightgray} 
\multicolumn{8}{l}{\textit{On the \la{KIT-ML dataset \cite{kit-ml}}.}}\\
\midrule[0.5pt]
     TEMOS \cite{TEMOS} & \textit{ECCV}'22  & $0.353^{\pm.006}$ & $0.561^{\pm.007}$ & $0.687^{\pm.005}$ & $3.717^{\pm.051}$ & $3.417^{\pm.019}$ & $10.84^{\pm.100}$ \\
     T2M-GPT \cite{t2m-gpt} & \textit{CVPR}'23 & $0.416^{\pm.006}$ & $0.627^{\pm.006}$ & $0.745^{\pm.006}$ & $0.514^{\pm.029}$ & $3.007^{\pm.023}$ & $10.86^{\pm.094}$\\
     ReMoDiffuse \cite{zhang2023remodiffuse}  & \textit{ICCV}'23 & $0.427^{\pm.014}$ & $0.641^{\pm.004}$ & $0.765^{\pm.055}$ & ${0.155}^{\pm.006}$ & $2.814^{\pm.012}$  & $10.80^{\pm.105}$\\
     MoMask \cite{Momask}  & \textit{CVPR}'24 & $0.433^{\pm.007}$ & $0.656^{\pm.005}$ & $0.781^{\pm.005}$ & $0.204^{\pm.011}$ & $2.779^{\pm.022}$ & - \\
     BAMM \cite{BAMM}  & \textit{ECCV}'24 & ${0.438}^{\pm.009}$ & ${0.661}^{\pm.009}$ & ${0.788}^{\pm.005}$ & $0.183^{\pm.013}$ & $2.723^{\pm.026}$ & $11.008^{\pm.094}$ \\
     MoGenTS \cite{MoGenTS} & \textit{NeurIPS}'24& $0.445^{\pm.006}$ & $0.671^{\pm.006}$ & $0.797^{\pm.005}$  & $\underline{0.143}^{\pm.004}$ & $2.711^{\pm.024}$ & $10.918^{\pm.090}$ \\
     Light-T2M \cite{light-t2m} & \textit{AAAI}'25 & $0.444^{\pm.006}$ & $0.670^{\pm.007}$ & $0.794^{\pm.005}$ & $0.161^{\pm.009}$ & $2.746^{\pm.016}$ & -\\ 
     IRG-MotionLLM \cite{li2025irg} & \textit{arXiv}'25 & $0.445^{\pm.005}$ & $0.681^{\pm.003}$ & $0.781^{\pm.004}$ & $0.432^{\pm.013}$ & $2.740^{\pm.017}$ & $11.115^{\pm.086}$ \\
     EnergyMoGen \cite{EnergyMoGen} & \textit{CVPR}'25 & $0.436^{\pm.006}$ & $0.651^{\pm.006}$ & $0.772^{\pm.006}$ & $0.495^{\pm.020}$ & $2.861^{\pm.020}$ & $11.06^{\pm.101}$ \\
     SALAD \cite{hong2025salad} & \textit{CVPR}'25 & $\underline{0.477}^{\pm.006}$ & $\bm{0.711}^{\pm.005}$ & $\bm{0.828}^{\pm.005}$ & $0.296^{\pm.012}$ & $\underline{2.585}^{\pm.016}$ & $11.097^{\pm.095}$ \\
    \midrule
     MotionHiFlow (ours) & - & $\bm{0.482}^{\pm.005}$ & $\underline{0.704}^{\pm.005}$ & $\underline{0.825}^{\pm.005}$ & $\bm{0.135}^{\pm.007}$ & $\bm{2.552}^{\pm.014}$ & $10.894^{\pm.117}$ \\
\bottomrule[1.5pt]
\end{tabular}
}
\vspace{-0.2cm}
\label{tab:main-results}

\end{center}
\end{table*}

In this section, we conduct extensive experiments on two {widely used} benchmark datasets. The results suggest that our MotionHiFlow consistently outperforms the current state-of-the-art methods quantitatively and qualitatively. Furthermore, ablation {studies demonstrate} the effectiveness of the key designs in our proposed framework. 

\subsection{Experimental Setup} \vspace{0.1cm}
\noindent
\textbf{Datasets.} We evaluate our MotionHiFlow framework on the widely used benchmark datasets: HumanML3D dataset~\cite{humanml3d} and KIT-ML dataset~\cite{kit-ml}.
The HumanML3D dataset \cite{humanml3d} comprises \(14,616\) human motions derived from \la{the AMASS \cite{amass} and HumanAct12 \cite{humanact12} } collections. Each motion is annotated with three distinct textual descriptions, resulting in \(44,970\) motion-text pairs in total. The KIT-ML dataset~\cite{kit-ml} contains \(3,911\) motions paired with \(6,278\) textual descriptions. Both datasets are split into training \(80\%\), validation \(5\%\), and test \(15\%\) sets.

\vspace{0.1cm}
\noindent
\textbf{Evaluation Metrics.}
We employ the same evaluation configuration as previous works \cite{humanml3d,t2m-gpt,Motiondiffuse,Momask}. To evaluate the semantic alignment between generated motions and input texts, we adopt \textit{R-Precision} and \textit{Multimodal Distance} as metrics, which is computed as the top-k recall precision and the multimodal distance between generated motions and input texts, respectively. 
We also employ the \textit{Fréchet Inception Distance (FID)} to measure the feature distributional similarity on the latent feature space between ground-truth (GT) and generated motions. 
Additionally, the \textit{Diversity} measures the variance across generated motion sequences, computed as the average Euclidean distance between 300 randomly sampled motion pairs. However, this metric typically shows similar values across methods and is thus less emphasized. 
Note that all metrics are calculated using pretrained a text encoder and motion encoder from T2M~\cite{humanml3d}.

\vspace{0.1cm}
\noindent
\textbf{Implementation Details.}
The Graph Convolutional Network (GCN) in our Motion VAE is adapted from 2s-AGCN~\cite{2s-agcn}. The encoder comprises two blocks, each integrating GCN and Temporal Convolutional Network (TCN) layers. This reduces the temporal dimension by a factor of $4$ (resulting in a sequence length of \(L/4\)) and pools the skeleton graph into \(j=6\) latent joints, representing the torso, pelvis, and four limbs (arms and legs). The weight of the augmentation loss (Eq.~\ref{eq:loss_aug}) is empirically set to $0.5$. For TMDiT, we employ nine blocks, with the first three using distinct parameters for the dual branches and the latter six sharing parameters. The latent dimension is set to 384, with 6 attention heads and a feed-forward dimension of 1536. The hierarchical flow matching comprises three flow layers, operating at \lh{scales} $r_k \in \{1/3, 2/3, 1\}$.

The MotionVAE is trained for 300,000 steps using AdamW~\cite{AdamW} with a batch size of 256 and an initial learning rate of \(2 \times 10^{-4}\). The TMDiT is subsequently trained for 200,000 steps using AdamW, with a batch size of 64 and an initial learning rate of \(2 \times 10^{-4}\). For both models, we apply a MultiStepLR scheduler, reducing the learning rate by a factor of 0.2 at 50\% and 75\% of the total training steps.

\subsection{Comparison with {State-of-the-art Methods}}
We compare our MotionHiFlow with various state-of-the-art methods, including VAE-based approaches~\cite{TEMOS}, autoregressive models~\cite{t2m-gpt,li2025irg}, diffusion-based models~\cite{Motiondiffuse,zhang2023remodiffuse,light-t2m,hong2025salad,EnergyMoGen} and discrete diffusion-like~\cite{Momask,MoGenTS} models.

\vspace{0.1cm}
\noindent
\textbf{Quantitative Results.}
Table \ref{tab:main-results} presents the quantitative comparisons between our MotionHiFlow and existing methods on the HumanML3D~\cite{humanml3d} and KIT-ML~\cite{kit-ml} datasets. Each experiment is repeated 20 times, with results reported along with a 95\% statistical confidence interval to ensure reliability. 
Table \ref{tab:main-results} shows that MotionHiFlow outperforms every baseline on both HumanML3D and KIT-ML. It achieves the highest R-Precision (0.563 / 0.482) and the lowest FID (0.032 / 0.135), evidencing tighter text–motion alignment and more realistic motion. Simultaneously, it records the smallest MultiModal Distance, indicating outputs that are both semantically coherent and varied. These results validate that our \lh{hierarchical} flow matching, joint-aware encoding, and diffusion transformer jointly set a new state of the art.




\begin{figure*}[htbp]
    \centering
    \includegraphics[width=\textwidth]{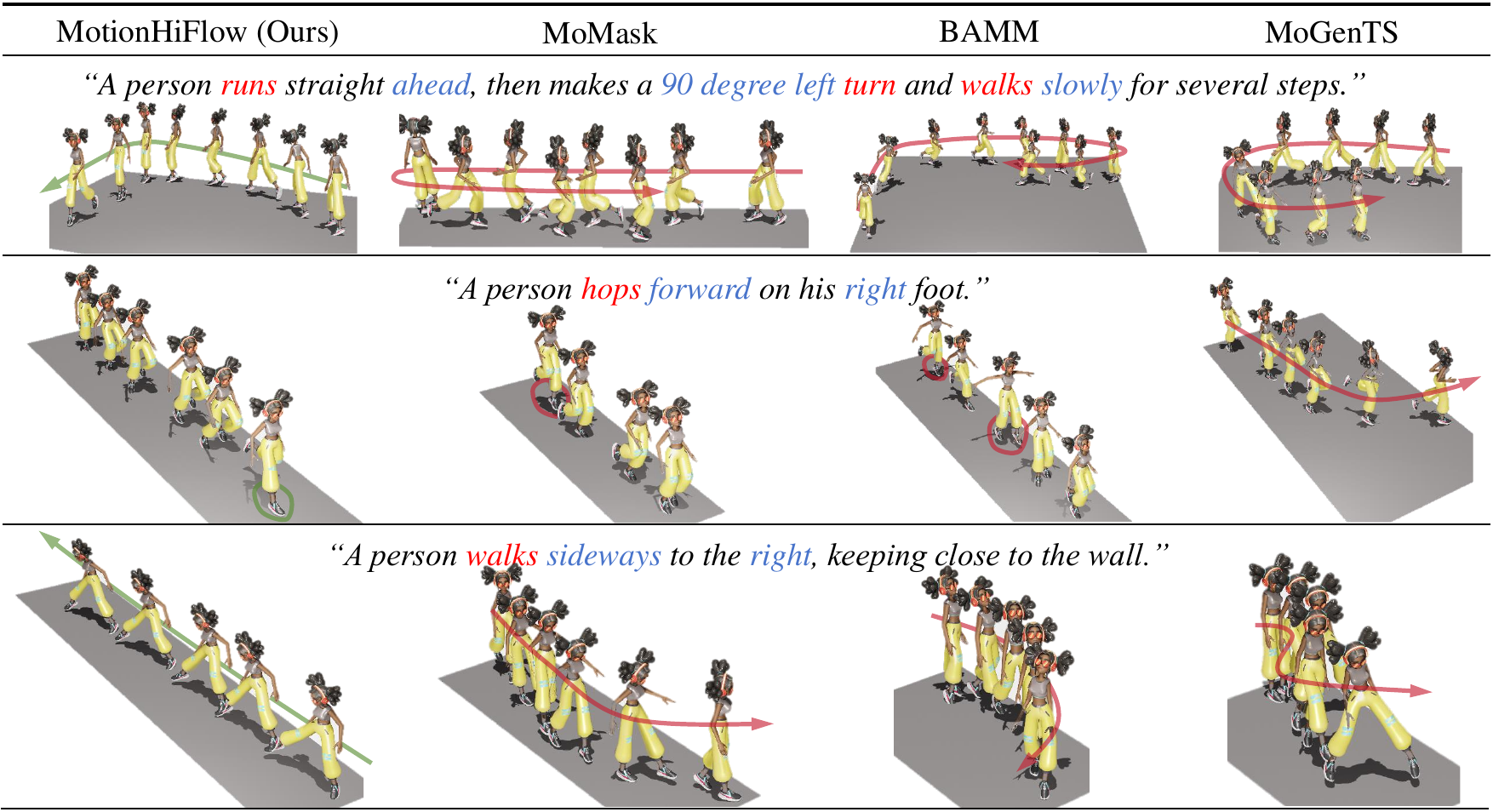}
    \caption{{Visual comparisons between different methods given three distinct text descriptions. Only key frames are displayed, with arrows indicating the character's movement direction. Green lines denote correct directions, while red lines indicate incorrect directions that do not match the text content. Refer to the demo video for complete motion clips and more visualization results.}}
    \label{fig:visualize}
\end{figure*}

\vspace{0.1cm}
\noindent
\textbf{Qualitative Results.}
Figure~\ref{fig:visualize} presents qualitative comparisons of our MotionHiFlow against prior methods, including Momask~\cite{Momask}, BAMM~\cite{BAMM}, and MoGenTS~\cite{MoGenTS}. As shown, existing approaches often misinterpret directional cues and specific limb movements. In contrast, MotionHiFlow produces motions with enhanced dynamic realism and strong alignment to textual descriptions, further validating the effectiveness of our hierarchical flow matching framework.

\subsection{Ablation Study}
In this subsection, we conduct ablation studies on the HumanML3D~\cite{humanml3d} dataset to comprehensively analyze the influence of different components in our MotionHiFlow, including the hierarchical flow matching and the architecture.

\begin{table}[!t]
  \footnotesize
  \centering
  \caption{Evaluation of system performance under varying scale settings. \lh{R@1 denotes top-1 retrieval precision.}}
  \vspace{-0.15cm}
  \label{tab:HT}
  {
  \begin{tabular*}{\linewidth}{@{}l@{\extracolsep{\fill}}c@{\extracolsep{\fill}}c@{\extracolsep{\fill}}c@{}}
    \toprule
     \lh{scales} $\{{r_k}\}$ & FID $\downarrow$ & R@1 $\uparrow$ & MM-Dist $\downarrow$ \\
    \midrule
    $[0.4]$ & $0.106^{\pm.004}$ & $0.561^{\pm.003}$ & ${2.717^{\pm.006}}$\\
    $[0.6]$ & $0.061^{\pm.003}$ & $0.556^{\pm.003}$ & $2.729^{\pm.006}$\\
    $[0.8]$ & $0.058^{\pm.002}$ & $0.559^{\pm.003}$ & $2.722^{\pm.005}$\\
      $[1]$ & $0.051^{\pm.002}$ & $0.556^{\pm.003}$ & $2.723^{\pm.006}$\\
    \midrule
    $[1/2, 1]$ & $0.038^{\pm.001}$ & $\bm{0.565}^{\pm.002}$ & $2.702^{\pm.006}$ \\
    \midrule
    $[1/3,2/3,1]$ & $\bm{0.032}^{\pm.002}$ & $0.563^{\pm.003}$ & $\bm{2.691}^{\pm.009}$ \\
    \midrule
    $[1/4,2/4,3/4, 1]$ & $0.035^{\pm.002}$ & $0.560^{\pm.003}$ & $2.693^{\pm.007}$ \\
    \bottomrule
  \end{tabular*}
  }
  \vspace{-0.3cm} 
\end{table}

\vspace{0.1cm}
\noindent
\textbf{Effectiveness of Hierarchical Flow Matching.}
In Table \ref{tab:HT}, we investigate the effect of hierarchical designs in our MotionHiFlow framework. Specifically, we test the system performance under varying numbers of hierarchical scales and different temporal scales. As can be seen: 1) even trained completely in a single coarse scale, our system can obtain MM-Dist ranging from 2.717 ($\{r_k\}=[0.4]$) to 2.729 ($\{r_k\}=[0.6]$), which means that a good semantic alignment is achieved; 2) performing hierarchical flow matching can improve both the FID and MM-Dist metrics. 

\begin{table}[!t]
  \footnotesize
  \centering 
  \vspace{-0.2cm}
  \caption{Evaluation of key components on the system performance.}
  \label{tab:Arch}
  {
  
  \begin{tabular*}{\linewidth}{@{}l@{\extracolsep{\fill}}c@{\extracolsep{\fill}}c@{\extracolsep{\fill}}c@{}}
  
    \toprule
    & FID $\downarrow$ & R@1$\uparrow$ & MM-Dist $\downarrow$\\
    \midrule
    Baseline & $0.074^{\pm.003}$ & $0.511^{\pm.003}$ & $3.043^{.008}$\\
    + TMDiT & $0.045^{\pm.003}$ & $0.557^{\pm.003}$ & $2.738^{\pm.007}$\\
    + topology-aware VAE & $\bm{0.032}^{\pm.002}$ & $\bm{0.563}^{\pm.003}$ & $\bm{2.691}^{\pm.009}$\\
    \bottomrule
  \end{tabular*}
  }
\end{table}

\vspace{0.1cm}
\noindent
\textbf{Analysis on the Architecture Designs.}
To evaluate the efficacy of the architectural components in our MotionHiFlow, we begin with the baseline model that employing a standard Transformer Encoder~\cite{Transformer} augmented with AdaLN~\cite{DiT}. We incrementally integrate the TMDiT module and the \lh{topology-aware Motion VAE}. As reported in Table~\ref{tab:Arch}, the incorporation of TMDiT yields a substantial improvement in the MM-Dist metric. We hypothesize that this enhancement stems from the word-level text encoding and the non-shared parameter strategy, validated through extensive experiments detailed in the supplementary material. Furthermore, the enhanced Motion VAE introduces spatial information, facilitating finer-grained motion generation.

\begin{figure}
    \centering
    \vspace{-2mm}
    \includegraphics[width=\linewidth]{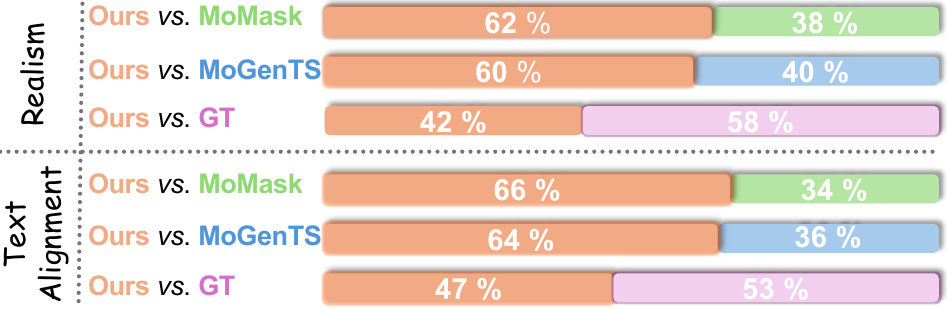}
    \caption{Results of a user study comparing the realism and text alignment of various methods, including our approach, with baseline competitors (MoMask, MoGenTS), and Ground Truth.}
    \label{fig:user_study}
\end{figure}

\vspace{0.1cm}
\noindent
\textbf{User Study.} We further conduct user study to compare the results generated by MotionHiFlow and previous methods including MoMask~\cite{Momask}, MoGenTS~\cite{MoGenTS}, and ground truth. We generate 100 motions for each pair of competitors using the text pool provided in HumanML3D test set and present the visualization results side-by-side. A total of 20 users are asked to vote which result is better from the aspect of realism and text alignment, respectively. The detailed results are presented in Figure~\ref{fig:user_study}. As shown, MotionHiFlow is preferred by users over MoMask and MoGenTS in both realism and text alignment, even with a 47\% chance of surpassing the ground truth in text alignment. 

\section{Conclusion}
\label{sec:conclusion}

In this work, we propose MotionHiFlow, a hierarchical flow matching framework for text-to-motion generation that progressively generates the motion from low to high temporal scales. By integrating a \lh{topology-aware} Motion VAE and Text-Motion Diffusion Transformer, our method generates motions with more superior semantic alignment, temporal coherence, and dynamic realism as compared with existing approaches, which is experimentally demonstrated on the HumanML3D~\cite{humanml3d} and KIT-ML~\cite{kit-ml} datasets.

\section*{Acknowledgements}
{
\small
This work was supported partially by the NSFC (62476296), Guangdong Natural Science Funds Project (2023B1515040025, 2022B1111010002, 2024A1111120017), Guangdong NSF for DistinguishedYoung Scholar (2022B1515020009), open research fund of Key Laboratory of Machine Intelligence and System Control, Ministry of Education (No. MISC-202407)
}




{
    \small
    \normalem
    \bibliographystyle{ieeenat_fullname}
    \bibliography{main}
}


\end{document}